\documentclass{article}

\usepackage{arxiv}

\usepackage[utf8]{inputenc} 
\usepackage[T1]{fontenc}    
\usepackage{hyperref}       
\usepackage{url}            
\usepackage{booktabs}       
\usepackage{amsfonts}       
\usepackage{nicefrac}       
\usepackage{microtype}      
\usepackage{lipsum}
\usepackage{xspace}
\newcommand{\eg}{e.g.\xspace}
\usepackage{graphicx}
\graphicspath{ {./images/} }
\usepackage{amsmath}

\title{EGSA-PT:Edge-Guided Spatial Attention with Progressive Training for Monocular Depth Estimation and Segmentation of Transparent Objects}

\author{
 Gbenga Omotara \\
  Vision-Guided and Intelligent Robotics Lab (ViGIR)\\
  University of Missouri\\
  Columbia, MO USA \\
  \texttt{goowfd@missouri.edu} \\
   \And
 Ramy Farag \\
  Vision-Guided and Intelligent Robotics Lab (ViGIR)\\
  University of Missouri\\
  Columbia, MO USA \\
  \texttt{rmf3mc@missouri.edu} \\
  \And
Seyed Mohamad Ali Tousi\\
  Vision-Guided and Intelligent Robotics Lab (ViGIR)\\
  University of Missouri\\
  Columbia, MO USA \\
  \texttt{stousi@missouri.edu} \\
   \And
G.N. DeSouza \\
  Vision-Guided and Intelligent Robotics Lab (ViGIR)\\
  University of Missouri\\
  Columbia, MO USA \\
  \texttt{desouzag@missouri.edu} \\
}

\begin{document}
\maketitle
\begin{abstract}

Transparent object perception remains a major challenge in computer vision research, as transparency confounds both depth estimation and semantic segmentation. Recent work has explored multi-task learning frameworks to improve robustness, yet negative cross-task interactions often hinder performance. In this work, we introduce Edge-Guided Spatial Attention (EGSA), a fusion mechanism designed to mitigate destructive interactions by incorporating boundary information into the fusion between semantic and geometric features. On both Syn-TODD and ClearPose benchmarks, EGSA consistently improved depth accuracy over the current state of the art method (MODEST), while preserving competitive segmentation performance, with the largest improvements appearing in transparent regions. Besides our fusion design, our second contribution is a multi-modal progressive training strategy, where learning transitions from edges derived from RGB images to edges derived from predicted depth images. This approach allows the system to bootstrap learning from the rich textures contained in RGB images, and then switch to more relevant geometric content in depth maps, while it eliminates the need  for ground-truth depth at training time. Together, these contributions highlight edge-guided fusion as a robust approach capable of improving transparent object perception. 
\end{abstract}


\section{Introduction}
\label{sec:intro}

Transparent object perception remains a challenging research problem in computer vision\cite{jiang2023robotic}, both in terms of recovering the 3D geometry of scenes with transparent objects\cite{cleargrasp_icra} or in understanding the semantics\cite{Mei_2022_CVPR} of these scenes. Devising robust solutions to these low-level vision tasks is essential for real world applications such as in robotics\cite{A4T_RAl} and autonomous driving\cite{AirSim_springer},\cite{md4all_ICCV}. In practice, failure to accurately perceive transparency can have serious consequences: a mobile robot may collide with a glass wall, or a manipulator may misjudge a transparent container during grasping. The difficulty arises because transparent objects violate the Lambertian assumption of light reflection. As a result, object boundaries become blurred and the continuity of surfaces is distorted and reflections are often misinterpreted by models. Conventional depth sensors also struggle to capture these objects reliably since their design likewise assumes Lambertian reflectance, often leading to reliance upon specialized sensors. Furthermore, the scarcity of high-quality annotated datasets for non-Lambertian objects makes training robust and reliable models for transparent object perception even more challenging. 

Recent approaches have sought to address the challenges of transparent object perception primarily through data-oriented strategies. For example, \cite{DAV2_Neurips}, \cite{Dpro_iclr}, \cite{MIDAS_TPAMI} curate large-scale datasets that aim to cover all possibilities and employ mixed dataset training to increase robustness across diverse scenarios including transparent ones. Other works tackle the problem by modifying the input data itself. The work done in \cite{Depth4Tom_ICCV} uses inpainting on transparent regions while \cite{Diffusion4Depth_ECCV} uses diffusion models to generate matte-equivalent training data. In addition, some researchers have explored specialized sensors \cite{Mei_2022_CVPR}, multi-view \cite{MVTrans_ICra}, or stereo setups \cite{SimNet_Corl}. While these methods show improvements and push the field forward, they primarily focus on improving the quality of the input data or extending modalities rather than crafting network modules that directly reason about transparent objects.

In parallel to these data-oriented approaches, another line of work focuses on designing network architectures that explicitly model the challenges of transparent object perception. For example, earlier works such as MVTrans\cite{MVTrans_ICra} and SimNet\cite{SimNet_Corl} adopted multi-task frameworks in multi-view and stereo settings, showing the value of simultaneously learning multiple perception tasks. However, their reliance on multi-view inputs limits their applicability in monocular scenarios. Building on this direction, MODEST\cite{MODEST_icra} introduced a monocular multi-tasking framework with a semantic-geometric fusion module which surpassed state-of-the art results on two challenging synthetic and real-world datasets. Their results demonstrate the benefits of the complementary cues across perception tasks for transparent objects. 

While multi-task frameworks benefit from the complementary cues between segmentation and depth, these cross-task interactions could also be detrimental. For example, segmentation may bias depth toward predicting nearly uniform values across an entire object, even when true depth varies significantly within that object. Similarly, depth may bias segmentation to group spatially adjacent regions with similar depth values into a single semantic category, even when they belong to different objects. Explicit boundary cues are therefore essential to prevent such misinterpretations, motivating our use of edge-guided fusion.

Beyond this cross-talk limitation, existing designs such as MODEST also rely heavily on channel attention, which we found can suppress a significant portion of the feature maps. In practice, this may either down-weight channels containing useful information or reveal a redundancy across many channels, leading to inefficient use of model capacity. This further reinforces the need for a redesigned fusion mechanism that incorporates boundary information that supports more reliable cross-task interaction.

\begin{figure*}[ht]
    \centering
    \includegraphics[width=0.9\linewidth]{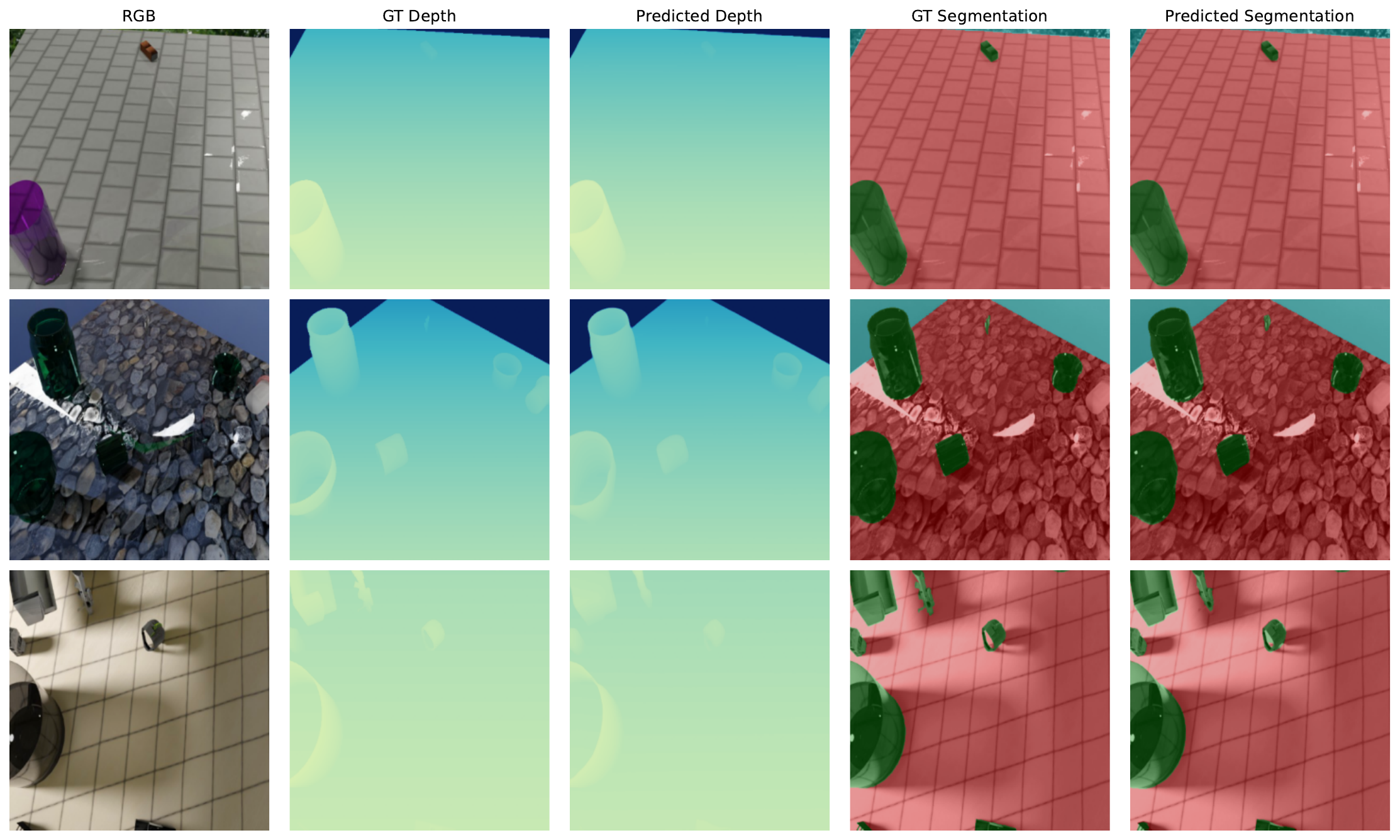}
    \caption{EGSA-PT predicts accurate depth and segmentation masks in the presence of transparent materials.}
    \label{fig:Model Predictions}
\end{figure*}

To this end, we propose an edge-guided redesign of the fusion mechanism that avoids the drawbacks of channel suppression. Our method introduces a gating mechanism for spatial attention by using edges as a guidance signal to incorporate boundary information. During training, these edges progressively shift from being generated from the input RGB images to being extracted from the model's predicted depth maps in a self-training manner, producing accurate predictions as in ~\ref{fig:Model Predictions}. In summary, our main contributions are:
\begin{itemize}
    \item  We introduce an edge-guided fusion mechanism that replaces channel attention with boundary aware spatial gating, allowing the model to better understand the bounds of objects.
    \item We propose a progressive training strategy that transitions from RGB-derived edges to depth-derived edges in a self-training manner, which improves stability and consistency.
\end{itemize}

\section{Related Work}
A growing body of work tackles transparent object perception by introducing high-quality synthetic and real-world datasets, since annotations for transparent objects are notoriously difficult to obtain in real-world contexts. Datasets such as Syn-TODD \cite{MVTrans_ICra}, ClearGrasp \cite{cleargrasp_icra}, ClearPose\cite{clearpose_ECCV}, Booster \cite{Booster_cvpr, Booster_tpami} and Trans10k \cite{Trans10k_eccv}  provide complementary resources across synthetic and real-world domains with diverse input modalities. For instance, Syn-TODD offers RGB-D, stereo and multi-view inputs with precise ground-truth annotations for depth, segmentation and bounding boxes under controlled rendering conditions. ClearGrasp likewise supplies RGB-D data with ground-truth depth, semantic segmentation, surface normals along with camera and object poses. These datasets make it possible to developing and benchmarking models tailored to the unique challenges presented by transparent objects.

Another line of work pursues data preprocessing strategies, where raw RGB inputs are transformed before being passed into a model. For example, \cite{Depth4Tom_ICCV} applies inpainting to transparent regions in images and generates pseudo-labels using a pre-trained monocular depth model\cite{DPT_ICCV}. These pseudo-lables are then used to fine-tune existing monocular\cite{DPT_ICCV}\cite{MIDAS_TPAMI} or stereo networks\cite{RAFT_3DV}. More recently, a diffusion-based method\cite{Diffusion4Depth_ECCV} generates images of transparent objects from their "matte" counterparts along with a depth map generated using an existing depth model\cite{DAV1_Cvpr} to ensure consistent scene geometry. This creates a dataset with "easy" (matte) and "hard" (transparent) image pairs with a shared depth map which is used to fine-tune existing depth models. These strategies improve training stability and generalization by providing more reliable supervision.

However, these techniques primarily act as front-end fixes. They enhance the quality of inputs or labels but leave the underlying model architectures unchanged and agnostic to the challenges of transparent objects. In contrast, other approaches tackle the problem at architectural design level, often through multi-tasking networks\cite{MVTrans_ICra}\cite{SimNet_Corl}\cite{MODEST_icra}. For example, MVTrans introduced a multi-view framework where partially overlapping images are back-projected into a 3D matching volume. This volume, combined with 2D features from a reference view, is then used to predict a low-resolution depth map, which supports downstream perception tasks. SimNet similarly proposed a stereo-matching strategy to learn robust low-level features, enabling sim-to-real transfer for downstream vision tasks. More recently, MODEST proposed a monocular multi-tasking network that explicitly fuses semantic and geometric features through iterative refinement, showing strong performance on both synthetic and real-world baselines.

While these methods demonstrate the promise of multi-task networks for transparent object perception, they differ in how they utilize inter-task interactions. Approaches such as SimNet and MVTrans do not explicitly exploit cross-task cues. MODEST, in contrast, highlights the benefits of such interactions but does not address their potential drawbacks. More broadly, prior work in multi-task learning has shown that jointly optimizing different tasks can lead to conflicting gradients, where one task pushes the other in a direction that hinders its learning.(\eg PCGrad\cite{PCGrad_Neurips}, GradNorm\cite{GradNorm}). We argue that multi-task networks for transparent object perception are particularly prone to such conflicts. For example, segmentation may encourage depth predictions to become overly uniform across an object, while depth may bias segmentation to merge adjacent objects that happen to be at similar distances. To counter these risks, we propose explicitly incorporating boundary information into the fusion process, guiding the interactions between both tasks in a way that preserves the benefits of multi-task learning while reducing harmful cross-talk.

\section{Method}
\begin{figure*}[ht]
    \centering
    \includegraphics[width=0.9\linewidth]{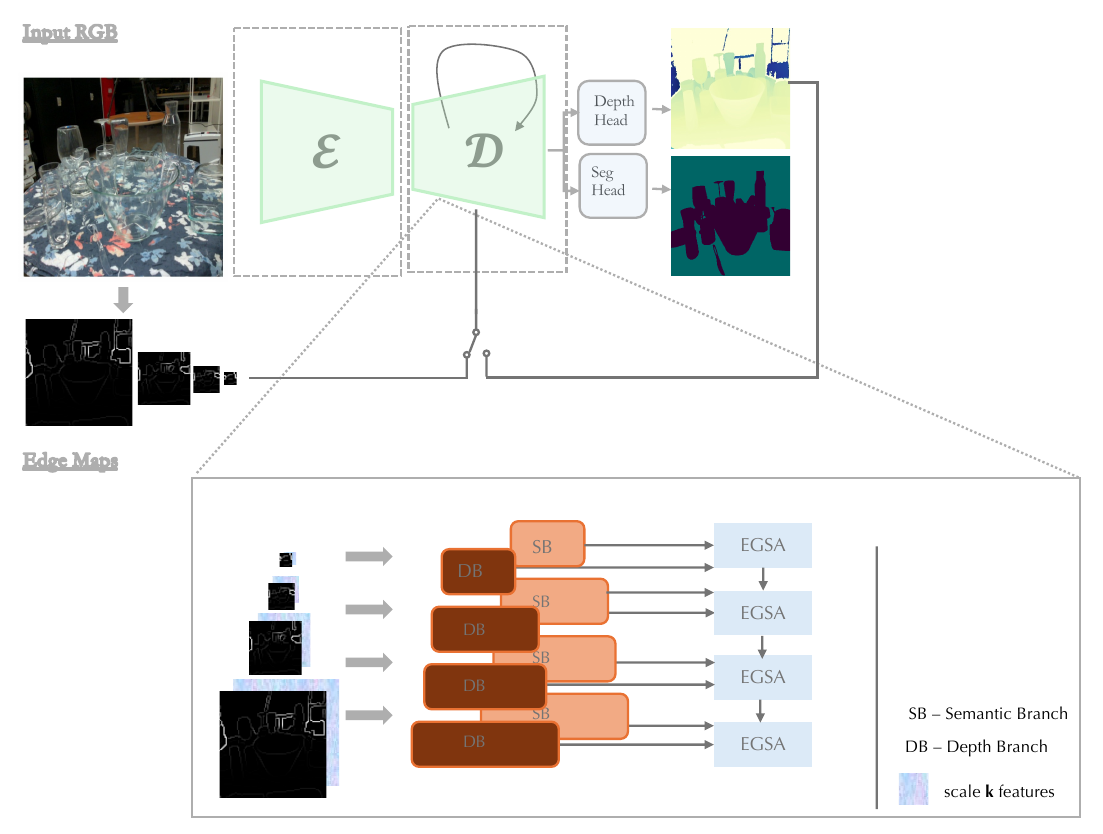}
    \caption{Overview of our proposed model architecture. A single RGB image is first encoded into multi-scale features with the encoder backbone. In parallel, multi-scale edges are injected into the EGSA block, together with multiscale features belonging to the depth and segmentation branch for fusion. The fused representations are then passed to the decoder for final depth and segmentation predictions.}
    \label{fig:model}
\end{figure*}

\subsection{Problem Statement and Overview}
Given a single RGB image $I \in \mathbb{R}^{3 \times H \times W}$ , the goal is to jointly predict a depth map $D \in \mathbb{R}^{H \times W}$ and a segmentation mask $S \in \mathbb{R}^{N \times H \times W}$ for transparent objects, where N denotes the number of semantic classes. Following prior work \cite{MODEST_icra}, we adopt a multi-task framework that learns a mapping function $f$ such that $(S,D) = f(I)$ as depicted in figure ~\ref{fig:model}. The baseline architecture in \cite{MODEST_icra}, is composed of three main components: a transformer-based encoder for extracting features, a reassemble module to construct multi-scale feature pyramids and an iterative fusion decoder that progressively refines the features. In this framework, the semantic and geometric branches are merged through a fusion module at each decoder stage, and predictions are iteratively refined in a coarse-to-fine manner. While we retain the encoder, reassemble module, and iterative refinement strategy of \cite{MODEST_icra}, our contributions lie in redesigning the fusion mechanism and introducing a progressive edge-guided training strategy. Specifically, we replace the channel-attention based fusion module with our edge-guided spatial attention mechanism, and we progressively transition from RGB-based edge guidance to model predicted depth edges during training. These modifications aim to enhance boundary sensitivity and reduce redundancy in feature utilization.

\subsection{Edge-Guided Fusion}
In \cite{MODEST_icra}, fusion between depth and segmentation features is facilitated by channel and spatial attention, where attention weights determine what and where to emphasize before combining task-specific features. While effective, this design relies on channel attention, which may suppress channels containing useful cues, and does not directly incorporate boundary information which is often useful for transparent object perception.

\begin{figure}[ht]
    \centering
    \includegraphics[width=0.9\linewidth]{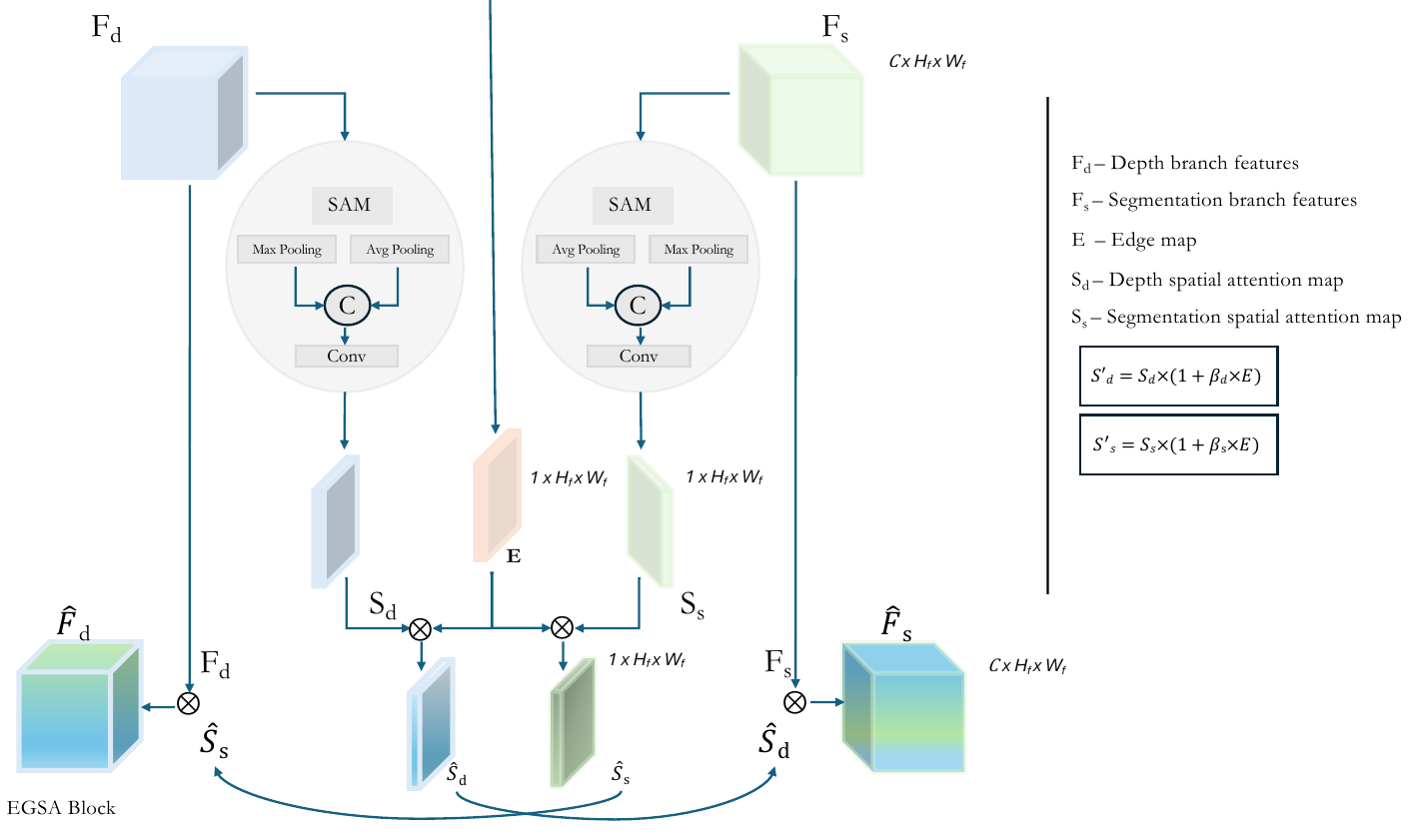}
    \caption{EGSA Block.}
    \label{fig:egsa_block}
\end{figure}

We therefore replace channel attention with an edge-guided modulation of spatial attention as illustrated in figure~\ref{fig:egsa_block}. Given encoder features at scale $k$, we denote  $F_d^k \in \mathbb{R}^{C \times H_k \times W_k}$ as the depth branch feature map and $F_s^k \in \mathbb{R}^{C \times H_k \times W_k}$ as the segmentation branch feature map. At the corresponding decoder scale, the spatial attention outputs derived from the segmentation and depth features $F_s^k$ and $F_d^k$  along with the edge map, namely $S^{k}_{s}, S^{k}_{d}, E^{k} \in \mathbb{R}^{1 \times H_k \times W_k}$, are used to produce gated attention signals as follows :

\begin{align}
gated\_seg^{k}   &= S^{k}_{s} \cdot \big(1 + \beta_{s2d}^{k} \cdot E^{k}\big), \\
gated\_depth^{k} &= S^{k}_{d} \cdot \big(1 + \beta_{d2s}^{k} \cdot E^{k}\big).
\end{align}

These gated attention maps are then applied to the corresponding features:

\begin{align}
\tilde{F}_s^k &= F_s^k \odot gated\_depth^{k}, \\
\tilde{F}_d^k &= F_d^k \odot gated\_seg^{k},
\end{align}

where $\odot$ denotes elementwise multiplication.

Here, $\beta^{k}_{s2d}$ and $\beta^{k}_{d2s}$ are learnable per-scale coefficients that adaptively control how strongly edges influence the spatial attention maps at each scale. The additive term "1" ensures that in the absence of edges at certain locations, the gating reduces to the original spatial attention, preventing the map from being suppressed entirely at those locations. This design allows the model to adaptively emphasize boundaries when edges are present, while retaining flexibility to rely solely on spatial attention in regions with no edges.

\subsection{Progressive Edge-Guided Training}
Training with self-generated edges from depth predictions introduces a chicken-and-egg problem: high-quality edges require accurate depth, yet, accurate depth benefits from our boundary-aware fusion guided by edges. Using predicted depth edges too early risks destabilizing training, as initial predictions are noisy and unreliable. On the other hand, relying solely on RGB-based edges risks overlooking task-specific geometric cues that emerge during learning.  

To resolve this dilemma, we adopt a \textbf{progressive edge-guided training strategy}. In the early epochs, edge maps are extracted from the RGB input image using a standard detector (Canny\cite{Canny}), providing a stable and consistent edge signal. After a warm-up period, we gradually transition to edges derived from the model's own depth predictions. Formally, at scale $k$:

\[
E^{k}(t) =
\begin{cases}
E^{k}_{\text{RGB}}, & t < T, \\
E^{k}_{\text{Depth}}, & t \geq T,
\end{cases}
\]

where $t$ is the epoch index and $T$ denotes the warm-up duration.  

This progressive shift functions as a form of curriculum learning: the model first learns from edges generated using the rgb-images and then it adapts to make use of its own geometry-aware cues once predictions become more accurate. As a result, the fusion mechanism benefits from stable supervision early on and increasingly task-specific boundary information later. By progressively shifting from image-based edges to self-predicted depth edges, the model learns to enforce boundaries that separate adjacent regions, thereby reducing harmful cross-task bias between depth and segmentation.

\subsection{Iterative Refinement}
We adopt the same iterative refinement strategy introduced in \cite{MODEST_icra}, where multi-scale features are updated and refined in a coarse-to-fine manner across three iterations. Lightweight gated units are used to transfer information between iterations. This design has proven effective for transparent object perception, and we therefore use it as the basis for our framework.

\subsection{Loss Functions}
Following \cite{MODEST_icra}, our model is trained in an end-to-end manner with two losses, one for depth and one for segmentation. The depth loss is defined as:
$$
L_{depth}=\|D-D^*\|_2 + \|\nabla{D}- \nabla{D^*}\|_1+\|N_D-N_{D^*}\|_1
$$
where the first term is the $\ell_2$ loss between the predicted depth $D$ and the ground-truth depth $D^*$, the second term is the $\ell_1$ loss between their gradients, and the last term is the $\ell_1$ loss between their surface normals.

The segmentation loss is the standard cross-entropy loss:
\begin{equation}
L_{\text{seg}} = \text{CE}(S, S^*),
\end{equation}
where $S$ and $S^*$ denote the predicted and ground-truth segmentation maps, respectively. 

The total loss is a weighted sum of both:
\begin{equation}
L = \alpha L_{\text{depth}} + \beta L_{\text{seg}},
\end{equation}
where $\alpha=1$ and $\beta=0.1$, following \cite{MODEST_icra}. 
These losses are applied at multiple scales and iterations of the multi-task decoder.

\section{Experiments}
\subsection{Datasets}
We evaluate our approach on two benchmark datasets for transparent object perception. Syn-TODD is a large scale synthetic dataset with photo-realistic images available in three modalities: RGB-D, stereo and multi-view rgb. Clearpose is a large-scale real world dataset, with 350k labeled RGB-D frames covering 63 household objects used in daily life under various lighting and occluding conditions. We adopt the standard training and testing splits as in \cite{MODEST_icra}. This combination allows us to assess performance under controlled synthetic scenarios and challenging real world cases. Because Syn-TODD is fully synthetic and photo-realistic, it provides controlled conditions with dense, noise-free ground truth for both depth and segmentation, making it suitable for detailed ablation studies. In contrast, ClearPose captures real-world complexity with occlusions and sensor noise, serving as a benchmark for robustness rather than fine-grained analysis.

\subsection{Implementation Details}
Our model uses a ViT-B encoder with an input resolution of $384 \times 384$. Training is conducted for $20$ epochs with a batch size of $4$. Optimization is performed using Adam, with separate learning rates of $1e-5$ for the encoder backbone and $3e-4$ for the decoder. Additionally, we use a progressive edge-guided training schedule where during the initial $T$ epochs, edges are extracted from the RGB inputs using the Canny operator. After a warm-up period, these RGB based edges are gradually replaced with edges derived from the model's predicted depth maps, enabling our fusion mechanism to incorporate geometry-aware boundaries.

\subsection{Evaluation Metrics}
For depth estimation, we report the Root Mean Squared Error (RMSE), Mean Absolute Error (MAE), absolute relative error with respect to the ground-truth (REL), and the percentage of pixels for which the maximum of the prediction-to-ground-truth and ground-truth-to-prediction ratios is below a given threshold, denoted as $\delta <1.05$, $\delta <1.10$, $\delta <1.25$. For segmentation, we report mean Average precision (mAP) and mean Intersection over Union (mIoU), with IoU $>0.5$ as the threshold in computing mAP.

\begin{table*}[ht]
\centering
\caption{Benchmark results on Syn-TODD. Our method improves depth accuracy while maintaining competitive segmentation performance. Bold indicates best performance.}
\label{tab:syntod_benchmark}
\begin{tabular}{l|ccc|ccc|cc}
\hline
Method & $\delta < 1.05$ & $\delta < 1.10$ & $\delta < 1.25$ & RMSE & MAE & REL & mAP & mIoU \\
\hline
MODEST (authors) & 65.28 & 78.23 & 93.05 & 0.07 & 0.052 & 0.068 & \textbf{97.82} & \textbf{92.84} \\
Ours (EGSA-RGB) & \textbf{68.38} & \textbf{80.16} & \textbf{94.45} & \textbf{0.068} & \textbf{0.044} & \textbf{0.059} & 97.59 & 92.39 \\
\hline
\end{tabular}
\end{table*}

\section{Results}

\subsection{Benchmark Results on Syn-TODD}
Table~\ref{tab:syntod_benchmark} reports benchmark results on Syn-TODD. Compared to MODEST, our edge-guided fusion (EGSA-RGB) achieves consistent improvements on depth estimation. The $\delta$-accuracies ($\delta < 1.05$, $\delta < 1.10$, $\delta < 1.25$) with respect to the ground truth improve by $+3.1\%$, $1.9\%$ and $1.4\%$ respectively indicating better handling of fine-grained depth variations with the greatest improvements in the strictest threshold. Error-based metrics (RMSE, MAE, REL) also decrease, confirming improved geometric accuracy. Segmentation performance (mAP, mIou) remains competitive, with less than $0.5\%$ variation relative to MODEST. Together, these results show that our edge guidance improves local consistency in depth predictions while preserving segmentation quality. 
\begin{figure*}[ht]
    \centering
    \includegraphics[width=0.9\linewidth]{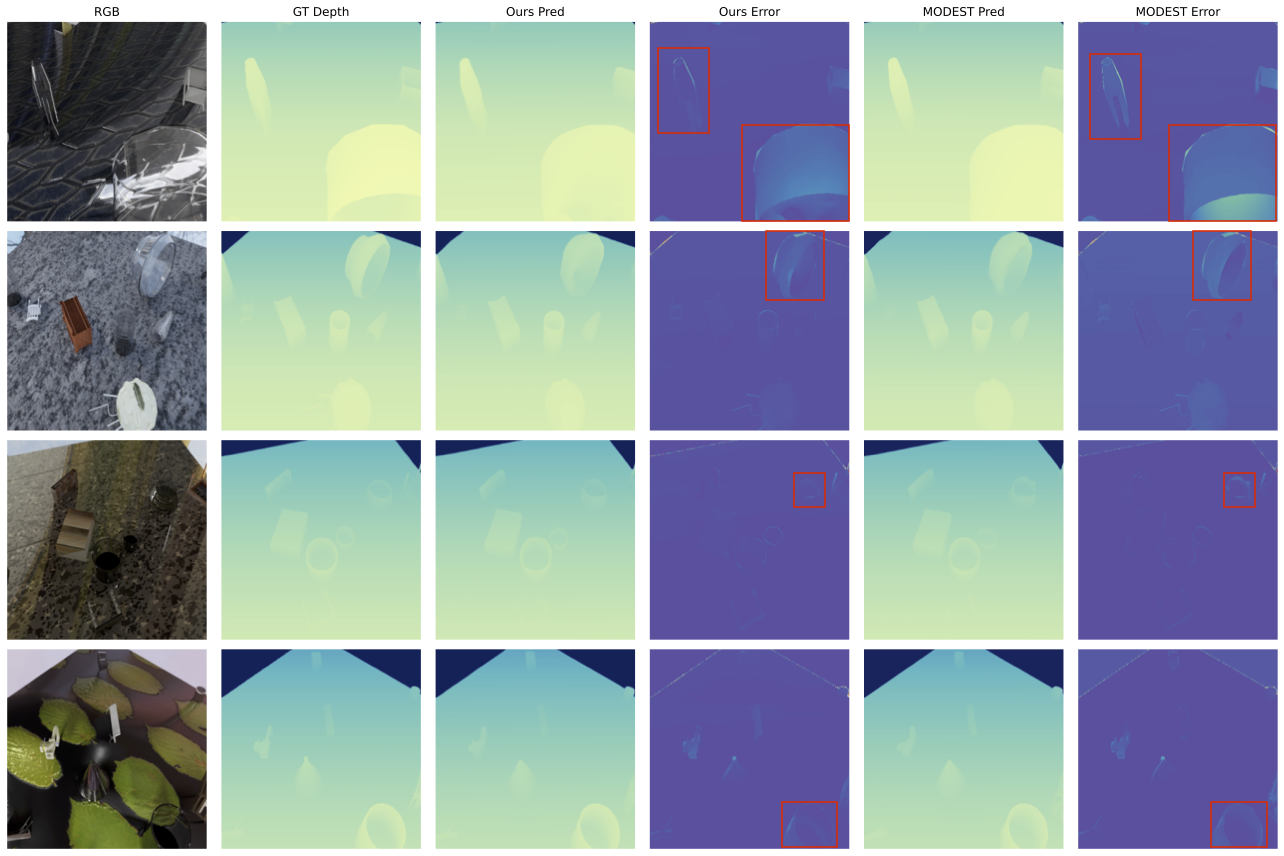}
    \caption{Qualitative comparison on Syn-TODD. Rectangles highlight transparent regions where our method reduces depth error compared to MODEST. Unlike MODEST, which often over-smooths depth, our edge-guided fusion preserves sharper transitions. This demonstrates the utility of boundary-aware fusion in precisely the regions where transparent perception is challenging. } 
    \label{fig:qualitative_comparison}
\end{figure*}

\subsection{Transparent only Analysis}
To specifically measure improvements on transparent regions, we report $\delta$-accuracies restricted to pixels belonging to transparent objects alone(Table ~\ref{tab:delta_transparent}). Our model achieves consistent gains over MODEST, with $+0.41\%$, $+1.04\%$ and $+1.08\%$ improvements on  $\delta < 1.05$, $\delta < 1.10$ and $\delta < 1.25$ respectively. Although these improvements are numerically small, they occur in challenging regions, where transparency amplifies negative cross-talk between geometry and semantics, making them especially meaningful. This consistent advantage highlights that edge-guided fusion delivers in the most challenging cases.

Figure \ref{fig:qualitative_comparison} illustrates qualitative comparisons between our method and MODEST on  transparent objects that are difficult to perceive even for the human eye. Significant regions with high residual errors are highlighted for the reader. In these areas, MODEST tends to over-smooth depth  or bleed object boundaries. In contrast, our edge-guided fusion produces sharper transitions and lower residual error, especially around object boundaries.  These visual improvements are consistent with the results observed in Table~\ref{tab:delta_transparent}, highlighting benefits around transparent regions.

\begin{table}[ht]
    \centering
    \begin{tabular}{l|c|c|c}
    \toprule
      & $\delta < 1.05 (T)$ & $\delta < 1.10(T)$ & $\delta < 1.25(T)$ \\
    \midrule
    Ours   & \textbf{52.82\%} & \textbf{71.48\%} & \textbf{91.25\%} \\
    MODEST & 52.41\% & 70.44\% & 90.17\% \\
    \bottomrule
    \end{tabular}
    \caption{Comparison of accuracy on transparent regions (T) on Syn-TODD. }
    \label{tab:delta_transparent}
\end{table}

\begin{table*}[ht]
\centering
\caption{Ablation study on Syn-TODD: The impact of attention components and edge-guidance. Depth metrics consistently improve when channel attention (CA) is removed while spatial attention (SA) is retained or edges when edges are added.}
\label{tab:ablation_fusion}
\begin{tabular}{l|ccc|ccc|cc}
\hline
Method Variant & $\delta < 1.05$ & $\delta < 1.10$ & $\delta < 1.25$ & RMSE & MAE & REL & mAP & mIoU \\
\hline
\multicolumn{9}{l}{\textit{Baseline}} \\
MODEST (CA + SA) & 65.28 & 78.23 & 93.05 & 0.070 & 0.052 & 0.068 & \textbf{97.82} & \textbf{92.84} \\
\hline
\multicolumn{9}{l}{\textit{Removing one component}} \\
MODEST (CA) & 67.16 & 79.24 & 93.49 & 0.071 & 0.049 & 0.065 & 97.08 & 91.62 \\
MODEST (SA) & \textbf{68.73} & \textbf{80.27} & \textbf{94.79} & 0.069 & 0.047 & 0.060 & 97.39 & 91.94 \\
\hline
\multicolumn{9}{l}{\textit{Edge-guided variants}} \\
EGSA (CA + SA) & 68.09 & 79.78 & 94.08 & 0.068 & 0.046 & 0.060 & 97.71 & 92.62 \\
EGSA (SA) & 68.38 & 80.16 & 94.45 & \textbf{0.067} & \textbf{0.044} & \textbf{0.059} & 97.59 & 92.39 \\
\hline
\end{tabular}
\end{table*}

\subsection{Ablations on Fusion Design}
To better understand where the improvements originate, we conduct an ablation study on the semantic and geometric fusion module on Syn-TODD, summarized in Table~\ref{tab:ablation_fusion}.

We first analyze the effect of removing individual attention components to assess their individual contributions to overall performance. Retaining only channel attention degrades both depth and segmentation on some metrics. Conversely, removing channel attention while keeping only spatial attention produces higher depth accuracy ($68.7$ vs $65.3$ baseline in the strictest $\delta$ metric ) with a similar reduction in segmentation as in channel attention only. This suggests that channel attention may not contribute as much in this setting.

Next, we examine the impact of guiding the fusion module with edges. Introducing RGB edge guidance on top of the baseline improves depth performance over the baseline while preserving segmentation performance, showing that boundary cues help mitigate destructive cross-task interaction. 
The best balance is achieved when spatial attention is combined with edge guidance, with channel attention removed, which consistently improves depth accuracy and maintains competitive segmentation. 
Taken together, these results reveal a consistent pattern: (i) channel attention is less effective compared to spatial attention in dense prediction,  and (ii) edge-guided spatial attention is more effective, since it emphasizes boundaries without sacrificing accuracy in segmentation. These findings reinforce our central claim, which is that, multi-task frameworks for transparent object perception suffers from cross-task bias, and replacing channel attention with edge guidance on the spatial features offers a principled solution to this challenge.


\begin{table*}[!ht]
\centering
\caption{Benchmark results on ClearPose. Comparison of MODEST and our variants with different edge sources. All model variants show consistent improvements over MODEST, underscoring the effectiveness of edge-guided fusion.}
\label{tab:clearpose_results}
\begin{tabular}{l|ccc|ccc|cc}
\hline
Method & $\delta < 1.05$ & $\delta < 1.10$ & $\delta < 1.25$ & RMSE & MAE & REL & mAP & mIoU \\
\hline
MODEST (authors) & 37.33 & 62.49 & 85.38 & 0.36 & 0.17 & 0.13 & \textbf{98.98} & \textbf{90.98} \\
 Ours (RGB Edges) & \textbf{52.09} & \textbf{73.66} & 88.41 & 0.34 & \textbf{0.14} & \textbf{0.11} & 98.08 & 85.04 \\
Ours (Depth Edges) & 50.53 & 73.62 & \textbf{88.63} & \textbf{0.32} & \textbf{0.14} & \textbf{0.11} & 98.17 & 87.30 \\
Ours (Progressive RGB$\rightarrow$Depth) & 51.86 & 73.59 & 88.51 & 0.34 & 0.15 & 0.12 & 98.20 & 84.61 \\ 
\hline
\end{tabular}
\end{table*}

\subsection{Results on ClearPose}
We further evaluate our method on the large-scale ClearPose dataset, which provides real-world RGB-D images of transparent household objects under clutter and occlusion. We adopt the official training and testing split from clearpose and compare our approach against MODEST using their released checkpoint. We note that these results differ from those reported in the MODEST paper, since the evaluation subset chosen by the authors is unknown;here, we stick with the official ClearPose testing set for consistency.

As shown in Table~\ref{tab:clearpose_results}, our variants outperforms MODEST on depth prediction metrics. In particular, using rgb-derived edges yields the strongest results, improving the strictest delta accuracy $\delta <1.05$ by more than $14\%$ relative to MODEST while also reducing errors across all depth metrics. Our depth-derived edge method along with our progressive strategy also provides solid gains across all depth metrics with respect to MODEST, confirming the validity of the hybrid strategy. Overall, we see significant improvements on a challenging real-world benchmark, further validating the effectiveness of our edge-guided attention. These results highlight that integrating boundary cues is especially beneficial in noisy real-world conditions even if segmentation metrics remain competitive rather than dominant.

\subsection{Edge Source Comparison (RGB vs Depth vs Progressive}


\begin{table*}[!htbp]
\centering
\caption{Impact of edge source on Syn-TODD. Comparing RGB based edges to Depth based edges to our hybrid approach with progressive training from RGB to Depth}
\label{tab:edge_source}
\begin{tabular}{l|ccc|ccc|cc}
\hline
Edge Source & $\delta < 1.05$ & $\delta < 1.10$ & $\delta < 1.25$ & RMSE & MAE & REL & mAP & mIoU \\
\hline
RGB Edges & 68.38 & 80.16 & 94.45 & \textbf{0.068} & \textbf{0.044} & \textbf{0.059} & 97.59 & 92.39 \\
Depth Edges & 66.32 & 79.49 & 93.97 & 0.070 & 0.052 & 0.064 & \textbf{97.87} & \textbf{92.84} \\
Progressive (RGB $\rightarrow$ Depth) & \textbf{68.49} & \textbf{80.52} & \textbf{94.71} & 0.069 & 0.049 & 0.060 & 97.79 & 92.85 \\
\hline
\end{tabular}
\end{table*}

Table~\ref{tab:edge_source} analyzes how the choice of the edge source affects model performance. Our starting hypothesis was that edges provide a valuable inductive bias for guiding feature fusion, but different sources carry different trade-offs and advantages. RGB-derived edges capture rich textures which do well to separate object boundaries across the scene, but they may also introduce spurious edges from surface textures that do not correspond to the geometry of the scene. Depth-derived edges on the other hand suppress non-geometry related artifacts while emphasizing true geometric boundaries, but these are less commonly used since they rely directly on the target being learned.

The results show these trade-offs. RGB edges achieve the best overall depth accuracy, with improvements in the $\delta$ accuracies over depth derived edges. Conversely, depth derived edges produces stronger segmentation performance over RGB derived edges. This suggests that depth-based boundaries provide better cues for semantic separation even if they compromise depth accuracy in geometry.

To consolidate the strengths of both edge sources, we introduce a progressive training strategy that starts with RGB edges in the early stages of training and then transitions to depth-derived edges from the models own predictions. This hybrid approach improves $\delta$ accuracies beyond depth derived edges while maintaining competitive segmentation performance, effectively combining the advantages of both. Although RGB edges remain the strongest, our progressive training offers a balanced trade-off, reducing cross-task bias while avoiding dependence on depth ground-truth during training.

\subsection{Backbone Study: DINOv2 + EGSA}
To examine the generality of our approach, we replace the backbone feature extractor used in our baseline (ViT-B\cite{VIT}) with DINOv2\cite{DINOV2}, a recent vision transformer that provides stronger feature representations. This experiment is an exploratory study to investigate whether our edge-guided fusion remains effective when paired with a more powerful backbone. 

\begin{table*}[!htbp]
\centering
\caption{Results on Syn-TODD with DINOv2 backbone. Our edge-guided fusion generalizes to stronger backbones, showing further improvements with RGB, Depth, and Progressive edge sources.}, 
\label{tab:dinov2_results}
\begin{tabular}{l|ccc|ccc|cc}
\hline
Method (DINOv2 backbone) & $\delta < 1.05$ & $\delta < 1.10$ & $\delta < 1.25$ & RMSE & MAE & REL & mAP & mIoU \\
\hline
Ours (RGB Edges) & \textbf{71} & \textbf{82.18} & \textbf{95.6} & 0.061 & \textbf{0.043} & \textbf{0.052} & 97.93 & \textbf{93.16} \\
Ours (Depth Edges) & 70.73 & 82.09 & 94.88 & 0.065 & 0.047 & 0.058 & 97.57 & 92.58 \\
Ours (Progressive RGB$\rightarrow$Depth) & 70.43 & 81.68 & 94.47 & \textbf{0.059} & \textbf{0.043} & 0.057 & \textbf{97.96} & 93.00\\
\hline
\end{tabular}
\end{table*}

As shown in Table~\ref{tab:dinov2_results}, EGSA continues to improve depth accuracy while maintaining strong segmentation performance when built on DINOv2. With RGB-derived edges, we observe further gains over our baseline on all metrics. While depth-derived and progressive edges also yield competitive results. These results demonstrate that the benefits of EGSA are not tied to a specific backbone. Instead, they scale with stronger feature extractors, further reinforcing the robustness of our design.

\section{Conclusions}

We introduced Edge-Guided Spatial Attention (EGSA), a fusion mechanism that incorporates boundary information to mitigate negative cross-task interactions in depth estimation and semantic segmentation. On Syn-TODD and ClearPose, EGSA consistently improved depth accuracy while maintaining competitive segmentation, with the largest benefits appearing in transparent regions where task interference is most severe.

We further compared different edge sources and proposed a progressive strategy that transitions from RGB to depth-derived edges. This balances the strengths of both cues without requiring ground-truth depth during training. Taken together, our findings position edge-guided fusion and progressive edge training as robust design principles for transparent object perception, while also motivating future work on learning task-specific edge representations.


\bibliographystyle{unsrt}  
\bibliography{references}  

\begin{thebibliography}{10}

\bibitem{jiang2023robotic}
Jiaqi Jiang, Guanqun Cao, Jiankang Deng, Thanh-Toan Do, and Shan Luo.
\newblock Robotic perception of transparent objects: A review.
\newblock {\em arXiv preprint arXiv:2304.00157}, 2023.

\bibitem{cleargrasp_icra}
Shreeyak Sajjan, Matthew Moore, Mike Pan, Ganesh Nagaraja, Johnny Lee, Andy Zeng, and Shuran Song.
\newblock Clear grasp: 3d shape estimation of transparent objects for manipulation.
\newblock In {\em 2020 IEEE International Conference on Robotics and Automation (ICRA)}, pages 3634--3642, 2020.

\bibitem{Mei_2022_CVPR}
Haiyang Mei, Bo~Dong, Wen Dong, Jiaxi Yang, Seung-Hwan Baek, Felix Heide, Pieter Peers, Xiaopeng Wei, and Xin Yang.
\newblock Glass segmentation using intensity and spectral polarization cues.
\newblock In {\em Proceedings of the IEEE/CVF Conference on Computer Vision and Pattern Recognition (CVPR)}, pages 12622--12631, June 2022.

\bibitem{A4T_RAl}
Jiaqi Jiang, Guanqun Cao, Thanh-Toan Do, and Shan Luo.
\newblock A4t: Hierarchical affordance detection for transparent objects depth reconstruction and manipulation.
\newblock {\em IEEE Robotics and Automation Letters}, 7(4):9826--9833, 2022.

\bibitem{AirSim_springer}
Shital Shah, Debadeepta Dey, Chris Lovett, and Ashish Kapoor.
\newblock Airsim: High-fidelity visual and physical simulation for autonomous vehicles.
\newblock In Marco Hutter and Roland Siegwart, editors, {\em Field and Service Robotics}, pages 621--635, Cham, 2018. Springer International Publishing.

\bibitem{md4all_ICCV}
Stefano Gasperini, Nils Morbitzer, HyunJun Jung, Nassir Navab, and Federico Tombari.
\newblock Robust monocular depth estimation under challenging conditions.
\newblock In {\em 2023 IEEE/CVF International Conference on Computer Vision (ICCV)}, pages 8143--8152, 2023.

\bibitem{DAV2_Neurips}
Lihe Yang, Bingyi Kang, Zilong Huang, Zhen Zhao, Xiaogang Xu, Jiashi Feng, and Hengshuang Zhao.
\newblock Depth anything v2.
\newblock In A.~Globerson, L.~Mackey, D.~Belgrave, A.~Fan, U.~Paquet, J.~Tomczak, and C.~Zhang, editors, {\em Advances in Neural Information Processing Systems}, volume~37, pages 21875--21911. Curran Associates, Inc., 2024.

\bibitem{Dpro_iclr}
Aleksei Bochkovskii, Ama\"{e}l Delaunoy, Hugo Germain, Marcel Santos, Yichao Zhou, Stephan~R. Richter, and Vladlen Koltun.
\newblock Depth pro: Sharp monocular metric depth in less than a second.
\newblock In {\em International Conference on Learning Representations}, 2025.

\bibitem{MIDAS_TPAMI}
Ren\'{e} Ranftl, Katrin Lasinger, David Hafner, Konrad Schindler, and Vladlen Koltun.
\newblock Towards robust monocular depth estimation: Mixing datasets for zero-shot cross-dataset transfer.
\newblock {\em IEEE Transactions on Pattern Analysis and Machine Intelligence}, 44(3), 2022.

\bibitem{Depth4Tom_ICCV}
Alex Costanzino, Pierluigi~Zama Ramirez, Matteo Poggi, Fabio Tosi, Stefano Mattoccia, and Luigi Di~Stefano.
\newblock Learning depth estimation for transparent and mirror surfaces.
\newblock In {\em 2023 IEEE/CVF International Conference on Computer Vision (ICCV)}, pages 9210--9221, 2023.

\bibitem{Diffusion4Depth_ECCV}
Fabio Tosi, Pierluigi {Zama Ramirez}, and Matteo Poggi.
\newblock Diffusion models for monocular depth estimation: {Overcoming} challenging conditions.
\newblock In {\em European Conference on Computer Vision ({ECCV})}, 2024.

\bibitem{MVTrans_ICra}
Yi~Ru Wang, Yuchi Zhao, Haoping Xu, Sagi Eppel, Alán Aspuru-Guzik, Florian Shkurti, and Animesh Garg.
\newblock Mvtrans: Multi-view perception of transparent objects.
\newblock In {\em 2023 IEEE International Conference on Robotics and Automation (ICRA)}, pages 3771--3778, 2023.

\bibitem{SimNet_Corl}
Thomas Kollar, Michael Laskey, Kevin Stone, Brijen Thananjeyan, and Mark Tjersland.
\newblock Simnet: Enabling robust unknown object manipulation from pure synthetic data via stereo.
\newblock In Aleksandra Faust, David Hsu, and Gerhard Neumann, editors, {\em Proceedings of the 5th Conference on Robot Learning}, volume 164 of {\em Proceedings of Machine Learning Research}, pages 938--948. PMLR, 08--11 Nov 2022.

\bibitem{MODEST_icra}
Yuan Liang, Bailin Deng, Wenxi Liu, Jing Qin, and Shengfeng He.
\newblock Monocular depth estimation for glass walls with context: A new dataset and method.
\newblock {\em IEEE Transactions on Pattern Analysis and Machine Intelligence}, 45(12):15081--15097, 2023.

\bibitem{clearpose_ECCV}
Xiaotong Chen, Huijie Zhang, Zeren Yu, Anthony Opipari, and Odest~Chadwicke Jenkins.
\newblock Clearpose: Large-scale transparent object dataset and benchmark.
\newblock In {\em European Conference on Computer Vision}, 2022.

\bibitem{Booster_cvpr}
Pierluigi Zama~Ramirez, Fabio Tosi, Matteo Poggi, Samuele Salti, Luigi Di~Stefano, and Stefano Mattoccia.
\newblock Open challenges in deep stereo: the booster dataset.
\newblock In {\em Proceedings of the IEEE conference on computer vision and pattern recognition}, 2022.
\newblock CVPR.

\bibitem{Booster_tpami}
Pierluigi~Zama Ramirez, Alex Costanzino, Fabio Tosi, Matteo Poggi, Samuele Salti, Stefano Mattoccia, and Luigi~Di Stefano.
\newblock Booster: A benchmark for depth from images of specular and transparent surfaces.
\newblock {\em IEEE Transactions on Pattern Analysis and Machine Intelligence}, 46(1):85--102, 2024.

\bibitem{Trans10k_eccv}
Enze Xie, Wenjia Wang, Wenhai Wang, Mingyu Ding, Chunhua Shen, and Ping Luo.
\newblock Segmenting transparent objects in the wild.
\newblock {\em arXiv preprint arXiv:2003.13948}, 2020.

\bibitem{DPT_ICCV}
René Ranftl, Alexey Bochkovskiy, and Vladlen Koltun.
\newblock Vision transformers for dense prediction.
\newblock In {\em 2021 IEEE/CVF International Conference on Computer Vision (ICCV)}, pages 12159--12168, 2021.

\bibitem{RAFT_3DV}
Lahav Lipson, Zachary Teed, and Jia Deng.
\newblock Raft-stereo: Multilevel recurrent field transforms for stereo matching.
\newblock In {\em International Conference on 3D Vision (3DV)}, 2021.

\bibitem{DAV1_Cvpr}
Lihe Yang, Bingyi Kang, Zilong Huang, Xiaogang Xu, Jiashi Feng, and Hengshuang Zhao.
\newblock Depth anything: Unleashing the power of large-scale unlabeled data.
\newblock In {\em CVPR}, 2024.

\bibitem{PCGrad_Neurips}
Tianhe Yu, Saurabh Kumar, Abhishek Gupta, Sergey Levine, Karol Hausman, and Chelsea Finn.
\newblock Gradient surgery for multi-task learning.
\newblock {\em arXiv preprint arXiv:2001.06782}, 2020.

\bibitem{GradNorm}
Zhao Chen, Vijay Badrinarayanan, Chen-Yu Lee, and Andrew Rabinovich.
\newblock Gradnorm: Gradient normalization for adaptive loss balancing in deep multitask networks, 2018.

\bibitem{Canny}
John Canny.
\newblock A computational approach to edge detection.
\newblock {\em IEEE Transactions on Pattern Analysis and Machine Intelligence}, PAMI-8(6):679--698, 1986.

\bibitem{VIT}
Alexey Dosovitskiy, Lucas Beyer, Alexander Kolesnikov, Dirk Weissenborn, Xiaohua Zhai, Thomas Unterthiner, Mostafa Dehghani, Matthias Minderer, Georg Heigold, Sylvain Gelly, et~al.
\newblock An image is worth 16x16 words: Transformers for image recognition at scale.
\newblock {\em arXiv preprint arXiv:2010.11929}, 2020.

\bibitem{DINOV2}
Maxime Oquab, Timothée Darcet, Theo Moutakanni, Huy~V. Vo, Marc Szafraniec, Vasil Khalidov, Pierre Fernandez, Daniel Haziza, Francisco Massa, Alaaeldin El-Nouby, Russell Howes, Po-Yao Huang, Hu~Xu, Vasu Sharma, Shang-Wen Li, Wojciech Galuba, Mike Rabbat, Mido Assran, Nicolas Ballas, Gabriel Synnaeve, Ishan Misra, Herve Jegou, Julien Mairal, Patrick Labatut, Armand Joulin, and Piotr Bojanowski.
\newblock Dinov2: Learning robust visual features without supervision, 2023.

\end{thebibliography}


\end{document}